# Generative AI trial for nonviolent communication mediation

Takeshi Kato[1][*]

[1] Hitachi Kyoto University Laboratory, Open Innovation Institute, Kyoto University, Kyoto, Japan

* Corresponding author

E-mail: kato.takeshi.3u@kyoto-u.ac.jp (TK)

# Abstract

Aiming for a mixbiotic society that combines freedom and solidarity among people with diverse values, I focused on nonviolent communication (NVC) that enables compassionate giving in various situations of social division and conflict, and tried a generative AI for it. Specifically, ChatGPT was used in place of the traditional certified trainer to test the possibility of mediating (modifying) input sentences in four processes: observation, feelings, needs, and requests. The results indicate that there is potential for the application of generative AI, although not yet at a practical level. Suggested improvement guidelines included adding model responses, relearning revised responses, specifying appropriate terminology for each process, and re-asking for required information. The use of generative AI will be useful initially to assist certified trainers, to prepare for and review events and workshops, and in the future to support consensus building and cooperative behavior in digital democracy, platform cooperatives, and cyber-human social co-operating systems. It is hoped that the widespread use of NVC mediation using generative AI will lead to the early realization of a mixbiotic society.

# Introduction

Social divisions and conflicts have become a global problem [1, 2]. Divisions and conflicts are not only between nations, such as Ukraine and Russia or Taiwan and China, but also between individuals to families, communities, regions [3, 4].

To solve these social issues, the philosopher Deguchi proposes the concept of "mixbiotic society" that further advances the idea of a symbiotic society [5–7]. The "mixbiotic society" is a society in which individuals with freedom and diverse values mix and mingle in physical proximity to recognize their respective "fundamental incapability" each other and sublimate into solidarity. The "fundamental incapability" is that the individual "I" is incapable of any physical action alone, nor of complete control over others. The subject of mixbiotic society is not "Self as I," but "Self as WE" who are entrusted to each other. The "Self as WE" is oriented toward diversity as other, freedom as de-regularity, and solidarity as mutual aid.

According to sociologist Luhmann, a social system is an autopoietic system that forms emergent order through a network of processes of communication generation and disappearance [8, 9]. Autopoiesis is a theory of living systems proposed by biologists Maturana and Varela, a self-organizing system consisting of a recursive network of processes in which components produce components through interaction [10, 11]. From the perspective of autopoiesis, communication is crucial for a social system in which freedom and solidarity are in harmony.

Nonviolent communication (NVC), developed by psychologist Rosenberg, is a well-known communication method for resolving social divisions and conflicts [12-14]. NVC is a method that creates empathy to promote interpersonal harmony and future cooperation, and the process consists of four components: observation, feelings, needs, and requests. NVC and the mixbiotic society have several things in common: NVC's fundamental questions "What's alive in us?" and "What can we do to make life more wonderful?" are similar to the "fundamental incapability" and "Self as WE" in the mixbiotic society. The avoidance of evaluation based on fixed values in observation and feelings corresponds to diversity of values, the avoidance of "should" and "must" in needs and requests corresponds to de-regularity, and the connections that allow for compassionate giving correspond to solidarity and mutual aid. Therefore, NVC is a powerful tool in aiming for a mixbiotic society. However, NVC is not for everyone to use immediately. To express observation with objectivity, feelings without evaluation, needs without coercion, and requests for positive action, mediation is necessary to support them.

Based on the above, the purpose of this report is to apply generative AI, which has recently made remarkable progress, to NVC mediation as part of research toward a mixbiotic society. Until now, mediation has been performed by skilled and certified trainers, of which there are only over 800 worldwide. However, if generative AI can be utilized, it would hasten the spread of NVC and the realization of a mixbiotic society. Generative AI is artificial intelligence that can generate text, images, or other media in response to prompts [15, 16]. Notable generative AIs include ChatGPT [17], Bing [18], Bard [19], and Stability AI [20]. In the Methods and Results section, I will test the ability of ChatGPT to perform appropriate mediation by entering examples into prompts for each of the four processes: observation, feelings, needs, and requests. In the Discussion section, I discuss future issues and developments based on the results of trial.

## Methods and Results

In this section, I will attempt NVC mediation using ChatGPT. Specifically, for each of the four processes (observation, feelins, needs, and requests), I first set up the mediator role and example sentences in ChatGPT at the prompt. Then, the sentences about observation, feelings, needs, and requests are entered as exercises in the prompt, and the output results in a response from ChatGPT. The responses are then rated on their suitability as mediation (+: good, 0: neither good nor bad, −: bad). The example sentences and exercises are taken from the literature [13].

# Observeation

| Input | # Role |
|---|---|
| | I will now enter a sentence, and you will revise it into a sentence that objectively states an "observation" without evaluation or judgment, or a sentence that separates objective "observation" from evaluation or judgment. If objective information is missing, output your guessed sentence. Below are three example sentences that you can learn from. |
| | # Example sentences |
| | 1. Input: John is far too generous. Output: That is not an "observation." What actions did John take? |
| | 2. Input: Doug put off. Output: What did Doug put off? |
| | 3. Input: Without a balanced diet, she would lose her health. Output: Since she is not eating a balanced diet (observation), I think her health will deteriorate (evaluation). |
| | # Confirmation |
| | Okay, I'm going to input a sentence, are you ready? |
| Output | Yes, I'm ready! Please go ahead and input the sentence you would like me to revise into an objective observation or separate observation from evaluation. |
| Input A | John got angry with me yesterday for no reason. |
| Output A | Observation: John got angry with me yesterday. <br> Separation of Evaluation: He got angry for no reason (evaluation).    **0** |
| Input B | My father is a good person. |
| Output B | Observation: My father is a person. <br> (Note: The original sentence does not provide specific actions or behaviors to objectively evaluate whether the person is "good." Hence, the revised observation only states the fact that the person mentioned is a father.)    **+** |
| Input C | Janice works too much. |
| Output C | Observation: Janice works a lot. <br> (Note: The original sentence contains the subjective term "too much," which implies an evaluation or judgment. The revised observation removes the evaluative term and states the objective fact that Janice works a lot.)    **+** |
| Input D | My son often does not brush his teeth. |
| Output D | Observation: My son often does not brush his teeth.    **−** |
| Input E | Luke said yellow doesn't look good on me. |
| Output E | Observation: Luke said yellow doesn't look good on me. <br> (Note: The sentence presents a direct quote without adding any evaluation or judgment to the statement.)    **+** |

Outputs A and C are highly evaluated for pointing out the lack of objective information, and it is desirable to be able to specify "no reason" in Output A and "often" in Output D. Overall, the mediation of observation by ChatGPT can be said to be successful.

# Feelings

| Input | # Role<br>I will now enter a sentence, and you will revise it into a sentence a subjective statement of "feelings" without evaluation or judgment, or as a sentence that separates subjective "feelings" from evaluation or judgment (e.g., "I feel ~ because I think ~."). If information about feelings is missing, output a guessed sentence. Three example sentences are presented below for your reference.<br># Example sentences<br>1. Input: I feel I have no talent as a guitar player. Output: That is not a "feeling." To paraphrase a desirable sentence, I am disappointed in myself as a guitar player.<br>2. Input: I am so mad at them for breaking the contract. Output: I felt angry at them because I think such behavior is irresponsible.<br>3. Input: Mom will be disappointed if you don't eat all of your food. Output: Mom will be disappointed if you don't eat all your food because mom wants you to grow up strong and healthy.<br># Confirmation<br>Okay, I'm going to input a sentence, are you ready? | |
|---|---|---|
| Output | Yes, I'm ready! Please go ahead and input the sentence you would like me to revise into a subjective statement of feelings or separate feelings from evaluation or judgment. | |
| Input A | I get frustrated every time you leave your work papers on the conference room floor. | |
| Output A | Subjective Feeling: I get frustrated every time you leave your work papers on the conference room floor.<br>Separation of Evaluation: I feel frustrated when you leave your work papers on the conference room floor because it makes the space messy and disorganized. | + |
| Input B | It is unpleasant that you are late. | |
| Output B | Subjective Feeling: It is unpleasant that you are late.<br>Separation of Evaluation: I feel unpleasant when you are late because it disrupts our plans and creates inconvenience. | + |
| Input C | I feel I am misunderstood. | |
| Output C | Subjective Feeling: I feel I am misunderstood.<br>(Note: The sentence is already expressing a subjective feeling without any evaluation or judgment.) | − |
| Input D | When you raise your voice, I am afraid. | |
| Output D | Subjective Feeling: When you raise your voice, I am afraid.<br>(Note: The sentence already presents a subjective feeling without any | + |

| | evaluation or judgment.) | |
|---|---|---|
| Input E | I am hurt by the little things that other people say. | |
| Output E | Subjective Feeling: I am hurt by the little things that other people say. (Note: The sentence is expressing a subjective feeling without evaluation or judgment.) | 0 |

Outputs A and B are highly commendable in that the reason for feeling is inferred. In Output C, "misunderstood" is an action expression, not a frank expression of emotion; Output E is acceptable in that it expresses a feeling, but it would be preferable if the needs at the root of feeling were mentioned. Overall, ChatGPT is applicable to the mediation of feelings.

# Needs

| | |
|---|---|
| Input | # Role<br>I will now enter a sentence, and you will revise it into a sentence a specific statement of "needs" without feelings, or as a sentence that separates specific "needs" from feelings (e.g., "I need ~, so I feel ~."). If information about needs is missing, output a guessed sentence. Three example sentences are presented below for your reference.<br># Example sentences<br>1. Input: I was lonely because he didn't come last night. Output: I needed intimacy with him, so I was lonely.<br>2. Input: I feel happy that you won the award. Output: Your award makes me happy because I needed the recognition of your achievements.<br>3. Input: I have never met such a self-centered person as you. Output: I need you to respect me, so your actions frustrate me.<br># Confirmation<br>Okay, I'm going to input a sentence, are you ready? |
| Output | Yes, I'm ready! Please go ahead and input the sentence you would like me to revise into a specific statement of needs or separate needs from feelings. |
| Input A | I am thankful that you let me ride in your car. |
| Output A | Specific Statement of Needs: I am thankful for the opportunity to ride in your car.<br>Separation of Feelings: I am thankful that you let me ride in your car because I needed transportation. | + |
| Input B | I was hoping that the work would have progressed further by now, but it did not, and I am disappointed. |
| Output B | Specific Statement of Needs: I was hoping that the work would have progressed further by now, but it did not. | + |

| | Separation of Feelings: I am disappointed because I needed the work to make more progress. | |
|---|---|---|
| Input C | I am fed up with the meetings dragging on and on. | |
| Output C | Specific Statement of Needs: I am fed up with the meetings dragging on and on.<br>(Note: The sentence already presents a specific statement of needs without feelings or evaluation.) | − |
| Input D | The reason I am upset that my boss broke his promise is because I had other plans for the weekend. | |
| Output D | Specific Statement of Needs: The reason I am upset that my boss broke his promise is that I had other plans for the weekend.<br>Separation of Feelings: I had other plans for the weekend, and I needed my boss to keep his promise, so I am upset. | 0 |
| Input E | Those people do not respect law and order, and it is inexcusable. | |
| Output E | Specific Statement of Needs: Those people do not respect law and order.<br>Separation of Evaluation: Their behavior is inexcusable because I need them to adhere to law and order for a peaceful society. | + |

Outputs A, B, and E are highly commendable in that they go one step further to express needs in response to the input sentences, while Outputs C and D are not expressed as needs. Overall, there is some applicability to the mediation of needs.

# Requests

| Input | # Role<br>I will now enter a sentence, and you will revise it into a sentence a specific statement of "needs" without feelings, or as a sentence that separates specific "needs" from feelings (e.g., "I need ~, so I feel ~."). If information about needs is missing, output a guessed sentence. Three example sentences are presented below for your reference.<br># Example sentences<br>4.  Input: I was lonely because he didn't come last night. Output: I needed intimacy with him, so I was lonely.<br>5.  Input: I feel happy that you won the award. Output: Your award makes me happy because I needed the recognition of your achievements.<br>6.  Input: I have never met such a self-centered person as you. Output: I need you to respect me, so your actions frustrate me.<br># Confirmation<br>Okay, I'm going to input a sentence, are you ready? |
|---|---|
| Output | Yes, I'm ready! Please go ahead and input the sentence you would like me to revise into a specific statement of needs or separate needs from feelings. |

| Input A | I am thankful that you let me ride in your car. | |
|---------|------------------------------------------------|---|
| Output A | Specific Statement of Needs: I am thankful for the opportunity to ride in your car.<br>Separation of Feelings: I am thankful that you let me ride in your car because I needed transportation. | + |
| Input B | I was hoping that the work would have progressed further by now, but it did not, and I am disappointed. | |
| Output B | Specific Statement of Needs: I was hoping that the work would have progressed further by now, but it did not.<br>Separation of Feelings: I am disappointed because I needed the work to make more progress. | + |
| Input C | I am fed up with the meetings dragging on and on. | |
| Output C | Specific Statement of Needs: I am fed up with the meetings dragging on and on.<br>(Note: The sentence already presents a specific statement of needs without feelings or evaluation.) | − |
| Input D | The reason I am upset that my boss broke his promise is because I had other plans for the weekend. | |
| Output D | Specific Statement of Needs: The reason I am upset that my boss broke his promise is that I had other plans for the weekend.<br>Separation of Feelings: I had other plans for the weekend, and I needed my boss to keep his promise, so I am upset. | 0 |
| Input E | Those people do not respect law and order, and it is inexcusable. | |
| Output E | Specific Statement of Needs: Those people do not respect law and order.<br>Separation of Evaluation: Their behavior is inexcusable because I need them to adhere to law and order for a peaceful society. | + |

The mediation of requests were rated low because, with the exception of Output B, no specific information was given on the behavior desired by the other persons. For example, Output A requires specific actions to make the other person understand, Output C requires positive actions to take instead of negative actions as stop drinking, Output D requires specific actions to respect privacy, and Output E requires specific frequency of "more often" is necessary. In other words, if these pieces of information are missing, it will be necessary to ask again from ChatGPT.

# Discussion

This trial of NVC mediation by ChatGPT showed that, in general, there is potential for the application of generative AI, although it is not at a level where it can be put to practical use immediately. The following are guidelines for improvement derived from the overall trend of the trial results.

- Increase the number of example sentences and model responses in the role setting prompts of generatinve AI. This will allow the AI to learn and increase the rate of good answers.

- Point out the bad points to the low rated responses and have the AI re-answer them. This allows it to learn.

- Specify the terms to be used in the response sentence. For example, for feelings and needs, you may refer to the Feelings and Needs Inventory in the literature [21]. These constraints make the response sentence the desired wording.

- Set up the prompt to ask again if necessary information is missing.

- Train to help the AI learn good and bad responces by certified trainers.

- Training of certified trainers will help the AI learn good and bad answers.

- Mediate by certified trainers and generative AI together at events and workshops to identify issues with generative AI.

- Utilize training databases collected by certified trainers, if available.

A good way to use generated AI is not to use it independently at first, but to use it to assist certified trainers or to prepare for or review events and workshops. In the future, it is expected that generative AI will be used as mediators in various situations of social division and conflict. Since it will be available online, it will be useful, for example, to support consensus building and cooperative action in digital democracy [22], platform cooperatives [23], and cyber-human social co-operating systems [24]. It is hoped that the widespread use of NVC mediation using generative AI will lead to the early realization of a mixbiotic society where freedom and solidarity among people with diverse values are compatible.

# Acknowledgements

I would like to express my gratitude to Professor Yasuo Deguchi of the Graduate School of Letters, Kyoto University, who is the principal investigator of the Japan Society for the Promotion of Science "Toward Better 'Smart WE'— From East Asian Humanities and Social Sciences to a Value Multi-Layered Society" project, and to Hitachi Kyoto University Laboratory for their input as the persons in charge. This work was supported by JSPS Topic-Setting Program to Advance Cutting-Edge Humanities and Social Sciences Research Grant Number JPJS00122679495.

# References

1. Schwander H. Labor market dualization and insider–outsider divides: Why this new conflict matters. Political Studies Review. 2018; 17(1): 14–29. https://doi.org/10.1177/147892991879087

2. Atran S. Psychology of transnational terrorism and extreme political conflict. Annual Review of Psychology. 2021; 72: 471–501. https://doi.org/10.1146/annurev-psych-010419-050800

3. Fredman LA, Buhrmester MD, Gomez A, Fraser WT, Talaifar S, Brannon SM, et

al. Identity fusion, extreme pro-group behavior, and the path to defusion. Social and Personality Psychology Compass. 2015; 9(9): 468–480. https://doi.org/10.1111/spc3.12193

4.  McFarland S, Hackett J, Hamer K, Katzarska-Miller I, Malsch A, Reese G, et al. Global human identification and citizenship: A review of psychological studies. Supplement: Advances in Political Psychology. 2019; 40(S1): 141–171. https://doi.org/10.1111/pops.12572

5.  Deguchi, Y. After corona seen from "Self-as-We." In: Hitachi Kyoto University Laboratory, editors. Beyond smart life, Japanese ed. Tokyo, Japan: Nikkei Business Publications; 2020.

6.  Deguchi, Y. Entrustment and distribution of agency: Toward philosophical reactivation of East Asian true self. Proceedings of Selfhood, Otherness, and Cultivation: Phenomenology and Chinese Philosophy International Conference; 2019 March 18–20; Nat'l Chengchi Univ, Taiwan. Available from: http://andy2.ccstw.nccu.edu.tw/wp-content/uploads/2022/03/0318.pdf

7.  Deguchi Y. Self-as-We & its ethical implications [Internet]. [cited 2019 Nov 10]; Maison des Sciences Humaines of the Université libre de Bruxelles. Available from: https://msh.ulb.ac.be/en/video/self-as-we-its-ethical-implications-yasuo-deguchi

8.  Luhmann N. Social Systems. Redwood City: Stanford University Press; 1996.

9.  Kneer G, Nassehi A. Niklas Luhmanns Theorie sozialer Systeme. München, Germany: Wilhelm Fink Verlag; 1993.

10. Varela F, Maturana H, Uribe R. Autopoiesis: The organization of living systems, its characterization and a model. Biosystems. 1974; 5: 187–196. https://doi.org/10.1016/0303-2647(74)90031-8

11. Kawamoto H. Autopoiesis: The third Stage of systems theory. Japanese ed. Tokyo, Japan: Seidosha; 1995.

12. Rosenberg MB. Speak peace in a world of conflict: What you say next will change your world. Encinitas: PuddleDancer Press; 2005.

13. Rosenberg MB. Nonviolent communication: A language of life. 3rd ed. Encinitas: PuddleDancer Press; 2015.

14. THE CENTER FOR NONVIOLENT COMMUNICATION [Internet]. n.d. Available from: https://www.cnvc.org/

15. Griffith E, Cade Metz C. Anthropic, an A.I. start-up, is said to be close to adding $300 million. The New York Times [Internet]. [cited 2023 Jan 27]. Available from: https://www.nytimes.com/2023/01/27/technology/anthropic-ai-funding.html

16. Lanxon N, Bass D, Davalos J. A cheat sheet to AI buzzwords and their meanings. Bloomberg [Internet]. [cited 2023 Jun 9]. Available from: https://www.bloomberg.com/news/articles/2023-06-09/what-s-generative-ai-what-s-machine-learning-an-ai-cheat-sheet

17. OpenAI. Introducing ChatGPT [Internet]. n.d. Available from: https://openai.com/blog/chatgpt

18. Microsoft Bing. Welcome to the new Bing [Internet]. n.d. Available from: https://www.bing.com/search?form=MY0291&OCID=MY0291&q=Bing+AI&sh

owconv=1

19. Bard. Try Bard, an AI experiment by Google [Internet]. n.d. Available from: https://bard.google.com/

20. Stability AI. AI by the people for the people [Internet]. n.d. Available from: https://stability.ai/

21. INTRODUCTION TO NVC [Internet]. n.d. Available from: https://www.cnvc.org/learn/what-is-nvc

22. Congge U, Guillamón MD, Nurmandi A, Salahudin, Sihidi IT. Digital democracy: A systematic literature review. Front. Polit. Sci. 2023; 5: 972802. https://doi.org/10.3389/fpos.2023.972802

23. Scholz T. Platform cooperativism. Ehmsen S, Scharenberg A, editors. New York: Rosa Luxemburg Stiftung; 2016. Available from: https://rosalux.nyc/wp-content/uploads/2020/11/RLS-NYC_platformcoop.pdf

24. Kato T, Kudo Y, Miyakoshi J, Owa M, Asa Y, Numata T, et al. Social Co-OS: Cyber-human social co-operating system. IET Cyber-Physical Systems: Theory & Applications. 2022; 8(1): 1–14. https://doi.org/10.1049/cps2.12037